\documentclass[10pt,twocolumn,letterpaper]{article}

\usepackage{cvpr}
\usepackage{times}
\usepackage{epsfig}
\usepackage{graphicx}
\usepackage{amsmath}
\usepackage{amssymb}
\usepackage{multirow}

\usepackage[pagebackref=true,breaklinks=true,letterpaper=true,colorlinks,bookmarks=false]{hyperref}

\cvprfinalcopy 

\def\cvprPaperID{1600} 

\begin{document}

\title{Multi-Scale Video Frame-Synthesis Network with Transitive Consistency Loss}

\author{Zhe Hu\\
Hikvision\\
{\tt\small zhe.hu@hikvision.com}
\and
Yinglan Ma\\
Adobe\\
{\tt\small yingma@adobe.com}
\and
Lizhuang Ma\\
East China Normal University\\
{\tt\small lzma@sei.ecnu.edu.cn}
}

\maketitle

\begin{abstract}
Traditional approaches to interpolate/extrapolate
frames in a video sequence require accurate pixel
correspondences between images, e.g., using optical flow.
Their results stem on the accuracy of optical flow estimation, and could generate heavy artifacts when flow estimation failed.
Recently methods using auto-encoder has shown impressive progress, however they are usually trained for specific interpolation/extrapolation settings and lack of flexibility and generality for more applications.
Moreover, these models are usually heavy in terms of of model size which constrains the application on mobile devices.
In order to reduce these limitations, we propose a unified network to parameterize the interest frame position and therefore infer interpolate/extrapolate frames within the same framework. 
To achieve this, we introduce a transitive consistency loss to better regularize the network. 
We adopt a multi-scale structure for the network so that the parameters can be shared across multi-layers. 
Our approach avoids expensive global optimization
of optical flow methods, and is efficient and flexible for video interpolation/extrapolation applications.
Experimental results have shown that our method performs favorably against state-of-the-art methods.
\end{abstract}

\section{Introduction}


Video frame synthesis, including interpolation and extrapolation, is a classic problem in computer vision and has attracted much interest recently due to the trend of unsupervised/self-supervised learning of video representation.
Frame interpolation has been used in numerous applications such as temporal upsampling, frame rate conversion and view synthesis.
Frame extrapolation, on the other hand, is related to predicting motion and it is a critical problem for learning interaction with the physical world, e.g., robots, autonomous driving of cars and drones.

Video frame synthesis itself is a challenging problem in the exist of moving deformable objects, object occlusion, illumination change, camera movement, and etc.. 
Traditional solutions to frame synthesis first compute dense correspondences, mostly optical flow, and then render image via correspondence-based image warping. 
Such methods heavily rely on computationally expensive global optimization, due to inherent ambiguities in computing correspondences.

To avoid explicit estimation of dense correspondences, recent
methods formulate frame interpolation~\cite{niklaus-cvpr2017-video,niklaus-iccv2017-video} or extrapolation~\cite{finn-nips2016-unsupervised,xue-nips2016-visual,bert-nips2016-dynamic} as a convolution process and estimate the convolution using the neural networks.
To speed up the convolution process with spatially-varying kernels, a spatially-adaptive separable convolution
approach has been presented for video frame interpolation~\cite{niklaus-iccv2017-video}, and obtain impressive interpolation results . However, they still pose the correspondence estimation as a separate step for the video interpolation process.

Most of the work that bypass correspondence estimation treat interpolation problem alone or extrapolation problem alone in a specific setting, e.g., interpolating the frame right in the middle of two input frames, extrapolating next frame with a certain interval. This is because existing networks are specifically designed for a fixed temporal position prediction, and lacks of generality and flexibility. In fact, if we consider a continuous sequence of video frames in temporal domain, either interpolating or extrapolating frames based on a subset of frames can be viewed as a unified problem that predicts frames at certain positions, and therefore should be handled as a whole. This is naturally solved in optical-flow-based methods, in which interpolation/extrapolation are handled along flow directions. Thus, how to elegantly formulate and solve the problem with a unified neural network becomes a critical problem.

In this work, we propose a unified end-to-end network for frame synthesize (interpolate/extrapolate) by parameterizing the temporal position of the interest frame.
To our best knowledge, this is the first work to synthesize video frames, \textbf{interpolating/extrapolating at any time ratio}, in a single network (without re-training for other ratio settings). 
The proposed Multi-Scale Frame-Synthesis Network (MSFSN) progressively
reconstruct interest frames in a coarse-to-fine manner.
We introduce a technique by enforcing transitive properties to enable training interpolation/extrapolation tasks in a single network. 
Experimental results show that our network is smaller than autoencoder based methods, e.g.,~\cite{niklaus-cvpr2017-video,niklaus-iccv2017-video,liu-arxiv2017-videoframe}, while generating comparable reconstruction accuracy.
Our network architecture naturally enables parameter sharing across pyramid levels, since different pyramid levels share the same input/output format and purpose. 
By sharing parameters across levels, we obtain a compact and flexible model that more levels can be stacked during test to accommodate higher capacity of deeper networks.

\begin{figure*}[t]
\begin{center}
\begin{tabular}{c}
\includegraphics[width = 0.96\linewidth]{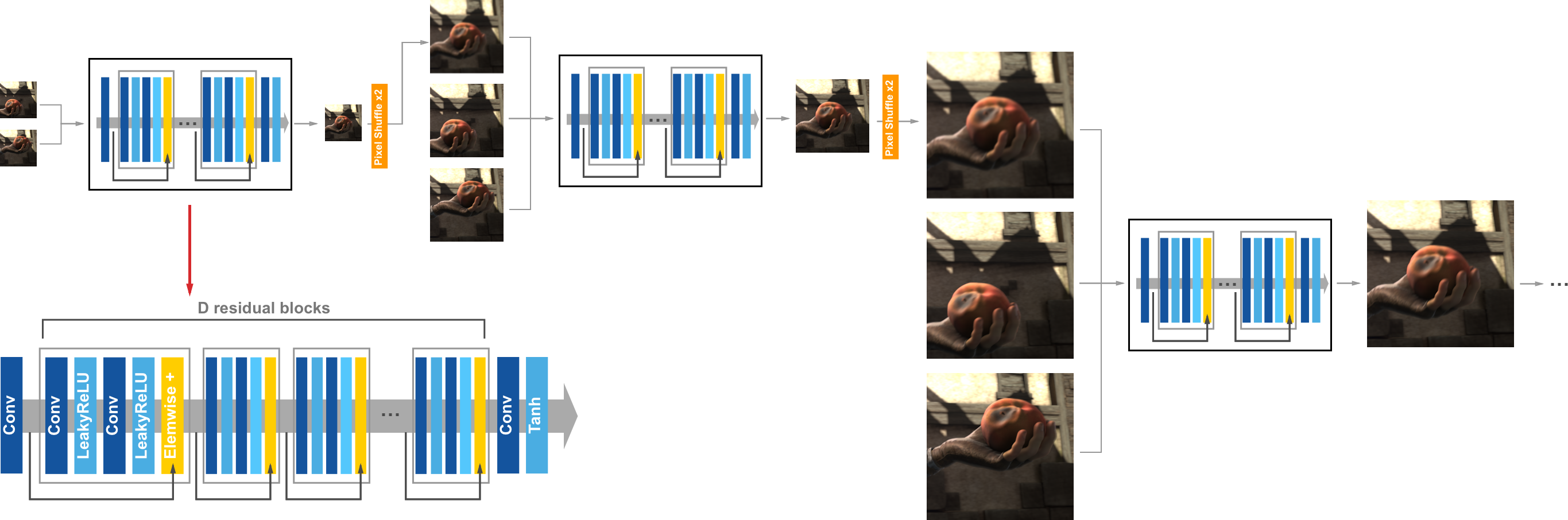} 
\end{tabular}
\caption{Example of our multi-scale network. Each level is a sub-network consisting of $D$ residual blocks~\cite{he-cvpr2016-deep} as shown in left bottom. Parameters are shared across pyramid levels, since different pyramid levels (except the coarsest level) share the same input/output format and purpose. We only show a network with 3 pyramid levels for simplicity.
}
\label{fig:network}
\end{center}
\vspace{-3mm}
\end{figure*}

\section{Related Work}
The proposed frame-synthesis GAN aims on interpolate/extrapolate video frames with a unified framework. We adopt a transitive consistency loss for leveraging the problems. Thus, we focus on reviewing work that are relevant to our problem and approach.

\textbf{Video frame interpolation/extrapolation}
Video frame interpolation is a classic topic in computer
vision and video processing. 
Traditional frame interpolation
methods estimate dense correspondence, mostly optical flow,
between input frames and then interpolate one or more
intermediate frames by warping the input images~\cite{baker-ijcv2011-database,werlberger2011optical}. 
The performance of these methods heavily stem on estimation accuracy of optical flow and require special processing
to reduce artifacts. 
Other than optical flow, approaches along view synthesis perspective have been explored~\cite{zitnick2004high,mahajan2009moving}. 
Meyer et al.~\cite{meyer-cvpr2015-phase} developed a phase-based interpolation
method that represents motion in the format of pixel phase shift and therefore render intermediate frames by modifying pixel phase.
This phase-based method often
produces impressive interpolation results, but could fail to preserve high-frequency details under large temporal changes.

Recent approaches formulate frame interpolation/extrapolation with a single convolution step that combines motion estimation and frame synthesis~\cite{xue-nips2016-visual,bert-nips2016-dynamic,niklaus-iccv2017-video,niklaus-cvpr2017-video,finn-nips2016-unsupervised}. 
These methods generate impressive results by estimating spatially-varying
kernels within the network and convolve them with input
frames to synthesize a new frame. Since these methods involves heavy convolution with large spatially variant kernels, they suffer from high computational load and memory for high-resolution videos.
Niklaus et al.~\cite{niklaus-iccv2017-video} propose a method to speed up the final convolution process via approximating 2D kernels with separable 1D kernel, and could process 1080p videos in one pass.
Recently, an approach has been proposed to
output dense voxel flows through a network and use them to generate interpolated frames~\cite{liu-arxiv2017-videoframe}. 
The generated 3D voxel flows encode the motion variance in temporal domain and generate intermediate frames via trilinear interpolation.

\textbf{Image reconstruction}
Our work is also inspired by the recent progress of network methods in image enhancement and reconstruction~\cite{gatys-arxiv2015-neural,johnson-eccv2016-perceptual,li2016precomputed,burger2012image,dong2016image,ledig-arxiv2016-photorealistic,lai-arxiv2017-deep,nah-arxiv2016-deep,dosovitskiy-nips2016-generating}. 
Pixel-wise loss functions such as MSE struggle to reconstruct high-frequency details, and therefore minimizing MSE usually lead to over-smooth results~\cite{mathieu-arxiv2015-deep,johnson-eccv2016-perceptual}.
Features extracted from a pre-trained VGG network instead of per-pixel error has been proposed to render visually pleasing images~\cite{bruna2015super,johnson-eccv2016-perceptual}. 
And perceptually more realistic results can be obtained via combining those with generative adversarial networks (GANs)~\cite{goodfellow-nips2014-generative} for image generation tasks~\cite{yu2016ultra,mathieu-arxiv2015-deep,li2016combining}.
%

%

\textbf{Transitive property}
Transitive property is a property of equivalence, and it has been used to regularize structured data for a long time in the literature. Forward-backward consistency check has proved to be efficient in depth estimation~\cite{hirschmuller-pami2008-stereo} and optical flow~\cite{sundaram-eccv2010-dense}. Language translation can also be improved via consistency check with back translation~\cite{brislin-jccp1970-back,he-nips2016-dual}.
More recently, higher-order cycle consistency has been used in
image-to-image translation~\cite{zhu-arxiv2017-unpaired}. In this work, we are introducing a
transitive consistency loss to facilitate training process for video representation.


\section{Frame-Synthesis Generative Adversarial Network}
Our goal is to learn a mapping function $G: S \times S \times T \rightarrow S$ in image domain $S$ and time domain $T$, for frame interpolation/extrapolation.
We denote a triplet $(x_{t_1}, x_{t_2}, t_p)$, where $x_{t_i} \in S, \quad i= 1,2$ represents an observed frame at timestamp $t_i$, and $t_p$ is the timestamp of our interest frame. Without loss of generality, we assume $t_1 < t_2$. 
We denote video frame distribution as $x_t \sim p_{data}$.
The mapping predicts the interest frame $y_{t_p} = G(x_{t_1}, x_{t_2}, t_p)$ at timestamp $t_p$, and $y_{t_p} \in S$.
If $t_1<t_p<t_2$, the problem is a frame interpolation problem.
If $t_p<t_1$ or $t_p>t_2$, the problem becomes a frame extrapolation problem. 

\subsection{Network Architecture}
We construct our network using a multi-scale structure~\cite{denton-nips2015-deep,ranjan-arxiv2016-optical,lai-arxiv2017-deep} as shown in Figure~\ref{fig:network}. Each level is a sub-network consisting of $D$ residual blocks~\cite{he-cvpr2016-deep}. We adopt a modified residual block by removing batch normalizations. We will discuss the effect of the block number in Section~\ref{sec:model_analysis}.

Our model takes a triplet $(x_{t_1}, x_{t_2}, t_p)$ as input and progressively predicts the interpolated/extrapolated frames from coarse levels to fine levels. Let $S$ and $y_{t_p}^s$ represent number of pyramid levels and the predicted frame at level $s, s \in \{1,2,\dots, S\}$ (from coarse to fine, level $S$ is the final reconstruction level).  Let $x_{t_1}$ and $x_{t_2}$ be $m \times n$ images that are multiples of $2^{S-1}$. 

At each pyramid level $s$, we feed downsampled frames $x_{t_1}^s, x_{t_2}^s$ of size $m/2^{S-s} \times n/2^{S-s}$ and an initial interpolated/extrapolated image $y_{t_p}^{s-1} \uparrow$ upsampled from previous level.  We predict the frame $y_{t_p}^s$ at level $s$ by passing input triplet $(x_{t_1}^s, x_{t_2}^s, y_{t_p}^{s-1} \uparrow)$ through our sub-network. In this work, we use a multi-scale structure of 4 levels and 3 sub-networks $\{N_s\}_{s=2}^4$ are $N_s :  (x_{t_1}^s, x_{t_2}^s, y_{t_p}^{s-1} \uparrow) =  y_{t_p}^{s}$. The sub-network at coarsest level $s=1$ takes $(x_{t_1}^1, x_{t_2}^1)$ as input, and is denoted as $N_1 :  (x_{t_1}^1, x_{t_2}^1) =  y_{t_p}^1$. \\

\textbf{Parameter sharing across pyramid levels}
The entire network is a cascade of sub-networks with the same structure
at levels except the coarsest level 1. 
We propose to share the network parameters across
those pyramid levels because the sub-networks at these levels share
the same structure and the task (i.e., predicting the interest frames with a lower-resolution version as input). 
As shown in Figure~\ref{fig:network}, we share the parameters of the sub-networks across
all the pyramid levels. 
As a result, the number of network
parameters is independent of the number of levels. 
We can use one single set of parameters to predict large motion by increasing the number of pyramid levels and this is discussed in Section~\ref{sec:model_analysis}.

\subsection{Loss Function}
Considering the supreme benefit of adversarial training on synthesizing realistic texture~\cite{ledig-arxiv2016-photorealistic,zhu-arxiv2017-unpaired}, we formulate our model in a GAN framework.
For the mapping function $G$, we introduce an adversarial discriminator $D$, where $D$ aims to distinguish between images and generated images by $G$~\cite{goodfellow-nips2014-generative}.
Our objective contains four terms: 1) pixel reconstruction loss; 2) feature reconstruction loss to encourage similarity in feature representation;  3) adversarial loss for matching the distribution of generated images to the data distribution in the target feature domain; 4) transitive consistency loss to enhance the mapping function $G$ with more constraints.

\begin{figure*}[tb]
\begin{center}
\begin{tabular}{ccc}
\includegraphics[width = 0.30\linewidth]{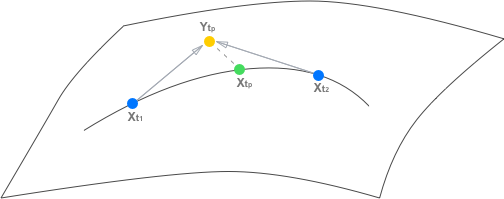}  & \hspace{-1mm}
\includegraphics[width = 0.30\linewidth]{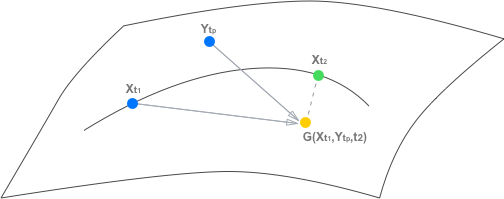}  & \hspace{-1mm}
\includegraphics[width = 0.30\linewidth]{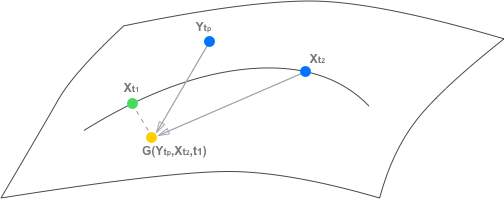}  \\
\end{tabular}
\caption{Example of transitive property. We aim to learn the frame synthesis mapping $G: S \times S \times T \rightarrow S$. Let $x_{t_1},x_{t_p},x_{t_2}$ denote three frames capturing the same scene in a video, and the mapping $y_{t_p} = G(x_{t_1}, x_{t_2}, t_p)$. The frames should lie on a manifold that represents the images of the scene, specifically a curve in the manifold that represents the frames in the video. To further regularize the mappings, we introduce a transitive consistency loss that captures the transitive mappings $G(y_{t_p},x_{t_2},t_1) \approx x_{t_1}$ and $G(x_{t_1},y_{t_p},t_2) \approx x_{t_2}$.
}
\label{fig:transitive}
\end{center}
\end{figure*} 

\textbf{Pixel reconstruction loss}
We adopt a per-pixel difference loss in $\ell_1$ norm as 
\begin{equation}
\mathcal{L}_{pix}(G) = \mathbb{E}_{p_{data}}[| G(x_{t_1}, x_{t_2}, t_p) -  y_{t_p} |_1],
\end{equation}
which is also suggested in~\cite{srivastava-arxiv2015-unsupervised,niklaus-iccv2017-video} to reduce blur effect rather than $\ell_2$ norm.

\textbf{Feature reconstruction loss}
Inspired by perceptual loss functions used in image reconstruction and style transfer networks~\cite{gatys-arxiv2015-neural,dosovitskiy-nips2016-generating,johnson-eccv2016-perceptual}, we use a feature reconstruction loss:
\begin{equation}
\mathcal{L}_{feat}(G) = \mathbb{E}_{p_{data}}[\| \phi(G(x_{t_1}, x_{t_2}, t_p)) -  \phi(y_{t_p}) \|_2].
\end{equation}
This loss is to encourage the feature of the output image to match that of the target image, and we use a similar loss network $\phi$ as~\cite{gatys-arxiv2015-neural,johnson-eccv2016-perceptual} based on 16-layer VGG network~\cite{simonyan-arxiv2014-very} pretrained on ImageNet~\cite{russakovsky-ijcv2015-imagenet}.

\textbf{Adversarial loss}
We apply the adversarial loss~\cite{goodfellow-nips2014-generative} to the mapping function $G$ and its discriminator $D$ as:
\begin{eqnarray}
\mathcal{L}_{GAN}(G,D) &=& \mathbb{E}_{p_{data}}[log_{x_{t_p} \ } D(x_{t_p})]  \\
&+& \mathbb{E}_{p_{data}}[log (1-D(G(x_{t_1}, x_{t_2}, t_p))], \nonumber
\end{eqnarray}
where $G$ tries to generate image $G(x_{t_1}, x_{t_2}, t_p)$ that looks similar to the frame $x_{t_p}$ at timestamp $t_p$, while $D$ aims to distinguish between generated image $G(x_{t_1}, x_{t_2}, t_p)$ and real sample $x_{t_p}$.
In other words, $G$ and $D$ are trained to optimize the objective in a minmax manner.
%

Adversarial training, in theory, can learn a discriminator $D$ to help the mapping function G
that produces outputs identically distributed as target
domains $S$ when $G$ is a stochastic function)~\cite{goodfellow-nips2014-generative}. 
However, with large enough network capacity, there are countless network with different parameter sets that can map a set of 
inputs to the target domain, with each of them mapping to a different image in
the target domain.
Thus, the adversarial loss alone cannot guarantee
that the learned function maps an input $(x_{t_1}, x_{t_2}, t_p)$ 
to the desired output $x_{t_p}$.


\textbf{Transitive consistency loss}
Since we aim at training the network for generic frame synthesis at any time ratio, special handling on the training loss is needed.
To further regularize the
mapping function, we argue that the learned mapping
function for frame interpolation/extrapolation should be transitive-consistent: given a mapping 
\begin{equation}
y_{t_p} = G(x_{t_1}, x_{t_2}, t_p),
\end{equation}
we should have transitive mappings for $x_{t_1}$ and $x_{t_2}$ giving $y_{t_p}$ as an input,
\begin{eqnarray}
G(y_{t_p},x_{t_2},t_1) &\approx& x_{t_1}, \nonumber \\
G(x_{t_1},y_{t_p},t_2) &\approx& x_{t_2}.
\end{eqnarray}
We call this \textit{transitive consistency}, and an illustration is shown in Figure~\ref{fig:transitive}. 
Therefore, we can formulate this behavior using a transitive consistency loss:
\begin{eqnarray}
\label{eq:tran_loss}
\mathcal{L}_{tran}(G) &=&  \mathbb{E}_{p_{data}}[| G(x_{t_1},y_{t_p},t_2) -  x_{t_2} |_1] \nonumber \\
&+& \mathbb{E}_{p_{data}}[| G(y_{t_p},x_{t_2},t_1) - x_{t_1} |_1].
\end{eqnarray}
%
%
In preliminary experiments, we also tried replacing the $\ell_1$
norm in this loss with an $\ell_2$, but did not observe
improved performance.

Another way to formulate transitive consistency is directly using the observed image $x_{t_p}$ instead of predicted image $y_{t_p}$ in~\eqref{eq:tran_loss}, as
\begin{eqnarray}
\label{eq:tran_loss2}
\mathcal{L}_{tran}(G) &=&  \mathbb{E}_{p_{data}}[| G(x_{t_1},x_{t_p},t_2) -  x_{t_2} |_1] \nonumber \\
&+& \mathbb{E}_{p_{data}}[| G(x_{t_p},x_{t_2},t_1) - x_{t_1} |_1].
\end{eqnarray}
This is similar to have all the triplet permutation in the same training batch, which is enforcing structured input for training.


\textbf{Objective function}
By combining adversarial loss and transitive consistency loss, our final objective function becomes:
\begin{eqnarray}
\mathcal{L}(G, D) =  \mathcal{L}_{pix}(G) + \lambda_{feat} \mathcal{L}_{feat}(G) + \\
\lambda_{GAN} \mathcal{L}_{GAN}(G,D) +  \lambda_{tran} \mathcal{L}_{tran}(G), \nonumber
\end{eqnarray} 
where $\lambda_{\ast}$ are the weights to balance different components.
Thus, our goal is to solve:
\begin{equation}
(G^{\star}, D^{\star} ) = \arg \min_{G} \max_{D} \mathcal{L}(G,D).
\end{equation} 

In Section~\ref{sec:model_analysis}, we compare our method
of the full objective, against ablations of the transitive consistency loss $\mathcal{L}_{tran}$ and its alternatives, and show that transitive consistency loss play a critical role
in obtaining high-quality results. 

\subsection{Implementation and Training Details}
In practice, instead of feeding the time stamp $t_p$ directly for mapping $G(x_{t_1}, x_{t_2}, t_p)$, we feed a ratio $r_p(t_1,t_2) = (t_p - t_1) / (t_2 - t_1)$ representing the relative temporal position comparing to $t_1$ and $t_2$.
Similarly, temporal ratios $(t_1 - t_p) / (t_2 - t_p)$ and $(t_2 - t_1) / (t_p - t_1)$ are used for mappings $G(y_{t_p},x_{t_2},t_1)$ and $G(x_{t_1},y_{t_p},t_2)$ respectively.
The ratio is formulated as a channel of the input image size and concatenated to the input images.

In the proposed MSFSN, all convolutional layers have
64 filters with the size of $5\times5$, and they are initialized using the Xavier initialization~\cite{glorot-icais2010-understanding}. 
We use the pixel-shuffle technique~\cite{shi-cvpr2016-real} in the upsampling layer.
We use the leaky rectified linear units (LeakyReLUs)~\cite{maas-icml2013-rectifier} with a negative slope of 0.2 as the non-linear activation function. 
For loss function, we set $\lambda_{feat} = 2*10^{-5}$, $\lambda_{GAN}=5*10^{-2}$ to balance with the pixel reconstruction loss $\mathcal{L}_{pix}$ according to~\cite{ledig-arxiv2016-photorealistic}.
The parameter $\lambda_{tran}$ is set to be $0.2$, and we determine the parameter via a coarse-to-fine cross-validation on a small validation dataset, with respect to the reconstruction accuracy and convergence speed.

The networks are trained using Adam optimization~\cite{kingma-arxiv2014-adam}
with $\beta_1 = 0.9$ and $\beta_2 = 0.999$. 
We use a batch size of 8, a patch size of 128
and 200 iterations per epoch for training. 
At each pass, three nearby frames are sampled from the videos with randomly selected intervals.
To guarantee that each training data contains enough information, we rule out smooth patches based on variance.
We use a learning rate of $1*10^{-4}$ for the initial generator training and decay it
during the adversarial training until the network converges. 
All our networks are trained on a single Nvidia P40 GPU.

We include various types of data augmentation during
training: 1) randomly rotate images by \{0, 90, 180, 270\} degrees; 2)  randomly flip images horizontally or vertically; 3) randomly crop patches of the same resolution of the training input; 4) adding additive Gaussian
noise sampled uniformly from $N(0, 0.1)$. 
During test, images are padded with mirror reflection such that their sizes are multiplies of $2^{S-1}$.

\section{Model Analysis}
\label{sec:model_analysis}
In this section, we first validate the contributions of different
components of the proposed network at an interpolation ratio 0.5 (interpolating middle in-between frame). We then discuss the
effect of multiple in-between frame interpolation. 
For validation experiments, we use the GOPRO dataset~\cite{nah-arxiv2016-deep}, in which the videos are captured using a GOPRO camera at 240 fps.
Since the videos are captured under significant camera movement, object motion and illumination change at a high sampling rate, the dataset well fits our purpose.
We split GOPRO dataset into training/test sets (2103/1111 frames) and train our model on the training set while evaluating on test set.

\begin{figure}[t]
\begin{center}
\begin{tabular}{ccc}
\includegraphics[width = 0.31\linewidth]{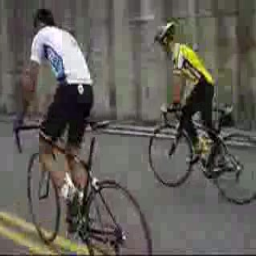} & \hspace{-4mm}
\includegraphics[width = 0.31\linewidth]{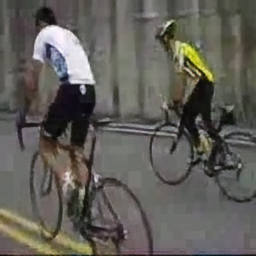} &\hspace{-4mm}
\includegraphics[width = 0.31\linewidth]{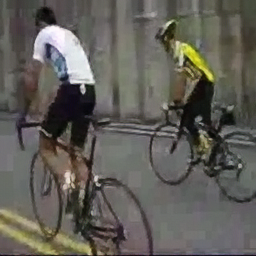}\\
\includegraphics[width = 0.31\linewidth]{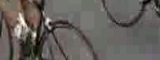} & \hspace{-4mm}
\includegraphics[width = 0.31\linewidth]{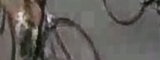} &\hspace{-4mm}
\includegraphics[width = 0.31\linewidth]{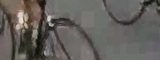}\\
(a) Ground truth & (b) Ours-Gen & (c) Ours \\
\end{tabular}
\caption{Example of frame interpolation on UCF-101 dataset~\cite{soomro-arxiv2012-ucf101}. We compare the proposed network with that trained without adversarial training (Ours-Gen).
}
\label{fig:gan_example}
\end{center}
\vspace{-3mm}
\end{figure}

\textbf{Adversarial training}
To validate the influence of the adversarial training, we train the proposed model with only generator (Ours-Gen) and compare it with the original model with the adversarial discriminator.  
We test this using the setting described in Section~\ref{sec:experiments} and show results in Table~\ref{tab:perf-interpolation}.
The proposed model does not outperform Ours-Gen in PSNR, but is able to render visually pleasing images. 
As shown in Figure~\ref{fig:gan_example}, our proposed method generate sharper images but the corresponding PSNR is slightly lower than those from Ours-Gen.
We note that our method removes some artifacts existed in the input images, e.g., the blocky artifacts in the example, thus is at a disadvantage in quantitative comparison.

\textbf{Pyramid depth}
Since the sub-networks in our model share same parameters, we can easily apply different numbers of pyramid levels during test with one trained model. 
We train our model of 4 pyramid levels on GOPRO training set and test the performance of different numbers of pyramid levels on GOPRO test dataset.
Moreover, we evaluate the results on different intervals: 1/2/3 frame intervals, meaning that the temporal distances between the interpolated frames and the pairs of input frames are 1/2/3 frames. The quantitative results are shown in Table~\ref{tab:pyr-depth}.
As shown in the table, stacking more levels in general leads to better performance (exception is 4-level structure with which the model is trained).
%
This is because long-range motion can be better handled by delving into lower downsampled images. 

\begin{table}
\centering
\begin{tabular}{c|ccc}
\hline
\multicolumn{1}{l|}{\multirow{2}{*}{\begin{tabular}[c]{@{}l@{}}Pyramid depth/ \\ Time interval\end{tabular}}} & \multicolumn{3}{c}{GOPRO (PSNR/SSIM)}                                          \\ \cline{2-4} 
\multicolumn{1}{l|}{}                                                                                         & \multicolumn{1}{c|}{1 frame}    & \multicolumn{1}{c|}{2 frames}   & 3 frames   \\ \hline
3 levels                                                                                                      & \multicolumn{1}{c|}{32.56/0.90} & \multicolumn{1}{c|}{31.98/0.88} & 31.39/0.81 \\
4 levels                                                                                                      & \multicolumn{1}{c|}{34.76/0.91} & \multicolumn{1}{c|}{34.24/0.90} & 33.56/0.85 \\
5 levels                                                                                                      & \multicolumn{1}{c|}{34.25/0.91} & \multicolumn{1}{c|}{33.71/0.90} & 33.04/0.84           \\
6 levels                                                                                                      & \multicolumn{1}{c|}{34.58/0.91} & \multicolumn{1}{c|}{34.04/0.90} & 33.35/0.84          \\
\end{tabular}
\caption{Quantitative evaluation on the number of pyramid levels. We train the model of 4 pyramid levels on GOPRO training set~\cite{nah-arxiv2016-deep} and evaluate the performance of different numbers of pyramid levels on GOPRO test set. The results are evaluated for frame interpolation at 1/2/3 frame intervals.}
\label{tab:pyr-depth}
\end{table}

\begin{table}
\centering
\begin{tabular}{c|c|c}
\hline
          & \#Params   & Size (MB) \\ \hline
DVF       & 13,413,440 & 53.7      \\ \hline
AdapSC    & 21,667,716 & 86.7      \\ \hline
Ours (12) & 9,878,617  & 39.5      \\ \hline
Ours (9)  & 7,419,481  & 29.7      \\ \hline
\end{tabular}
\caption{Comparisons on parameter number and model size. We compare the proposed method of sub-network depth 9 and 12 with DVF~\cite{liu-arxiv2017-videoframe} and AdapSC~\cite{niklaus-iccv2017-video}.}
\label{fig:param}
\end{table}

\textbf{Sub-network depth}
Each of the sub-network contains $D$ residual blocks and we explore the performance influence on the sub-network depth, i.e., number of residual blocks.
We train the proposed model of 4 pyramid levels with different depth, $D = 5,9,12$ at each level, and show the performance on interpolation task in Table~\ref{tab:net-depth}. 
The models are trained on GOPRO training set and evaluated on GOPRO test dataset.

In general, deeper networks perform better than shallower ones at the expense of increased training time and computational cost. 
We use $D = 9$ in our model for most of our experiments to compromise between accuracy and speed. 

\begin{table}[]
\centering
\resizebox{\linewidth}{!}{%
\begin{tabular}{c|l|ccc}
\hline
\multicolumn{1}{l|}{\multirow{2}{*}{\begin{tabular}[c]{@{}l@{}}Network depth/\\ Time interval\end{tabular}}} & \multicolumn{1}{c|}{\multirow{2}{*}{\#Parameters}} & \multicolumn{3}{c}{GOPRO (PSNR/SSIM)}                                    \\ \cline{3-5} 
\multicolumn{1}{l|}{}                                                                         & \multicolumn{1}{c|}{}                              & \multicolumn{1}{c|}{1 frame} & \multicolumn{1}{c|}{2 frames} & 3 frames \\ \hline
5   &     \multicolumn{1}{c|}{4,140,633}                                  & \multicolumn{1}{c|}{34.43/0.90}     & \multicolumn{1}{c|}{33.82/0.88}     & 33.13/0.81      \\
9   &     \multicolumn{1}{c|}{7,419,481}                                   & \multicolumn{1}{c|}{34.76/0.91}     & \multicolumn{1}{c|}{34.24/0.90}     &  33.56/0.85     \\
12  &    \multicolumn{1}{c|}{9,878,617}                                   & \multicolumn{1}{c|}{34.80/0.91}     & \multicolumn{1}{c|}{34.27/0.90}     &   33.65/0.85    \\
\end{tabular}}
\caption{Quantitative evaluation on the sub-network depth at each level. We build MSFSN with different sub-network depth by
varying the numbers of residual blocks. The results are evaluated for frame interpolation task on the UCF-101 dataset~\cite{soomro-arxiv2012-ucf101}.}
\label{tab:net-depth}
\end{table}

\begin{figure*}[t]
\begin{center}
\begin{tabular}{cc}
\includegraphics[width = 0.40\linewidth]{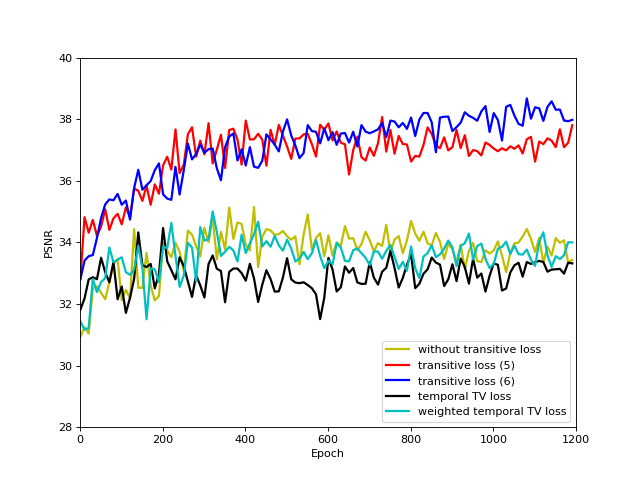} &
\includegraphics[width = 0.50\linewidth]{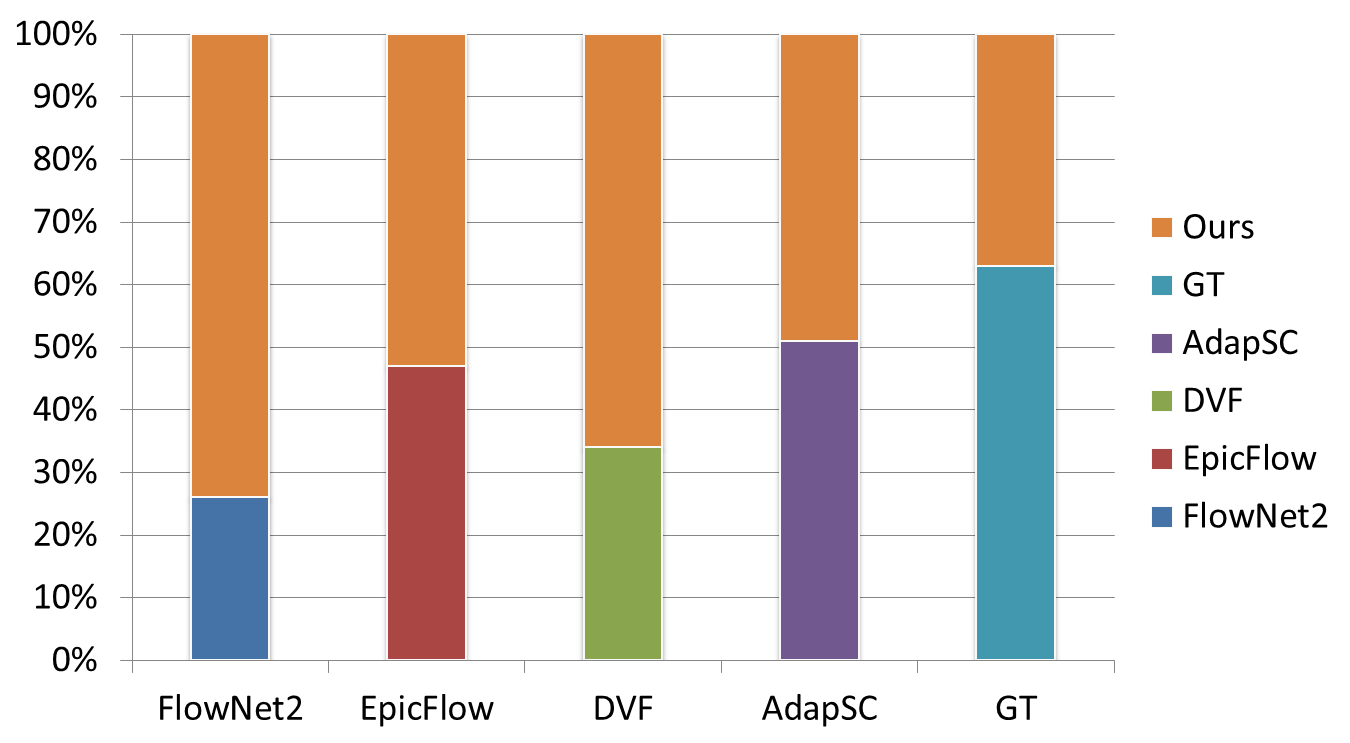} \\
(a) & (b) \\
\end{tabular}
\caption{(a) Training curves of transitive consistency loss and its alternatives in terms of PSNR value. (b) User study results on frame interpolation. We compare our method with state-of-the-art methods, FlowNet2~\cite{ilg-arxiv-2016-flownet}, EpicFlow~\cite{revaud-cvpr2015-epicflow}, DVF~\cite{liu-arxiv2017-videoframe}, AdapSC~\cite{niklaus-iccv2017-video} and the ground truth (GT).}
\end{center}
\label{fig:tran_comp}
\end{figure*}

\textbf{Transitive consistency loss.}
To validate the effectiveness of the transitive consistency loss,
we compare with the proposed network with alternatives, and show the training curve in terms of PSNR metric instead of loss curve, as the loss function has been changed when removing transitive consistency loss.
%
We conduct experiments by comparing following alternatives on a validation dataset under 1200 epochs (evaluate every 10 epochs): 1) no transitive consistency loss $\mathcal{L}_{tran}$, but increasing the weight for pixel reconstruction loss to 1.4, to match the loss level when using transitive consistency loss; 2) using transitive consistency loss in~\eqref{eq:tran_loss}; 3) using transitive consistency loss in~\eqref{eq:tran_loss2}; 4) using a temporal total variation (TV) loss 
\begin{equation}
\mathbb{E}_{p_{data}}[| G(x_{t_1},x_{t_2},t_p) -  x_{t_1} |_1] + \mathbb{E}_{p_{data}}[| G(x_{t_1},x_{t_2},t_p) - x_{t_2} |_1],
\end{equation} 
instead of $\mathcal{L}_{tran}$; 5) using a weighted temporal TV loss based on the temporal distance as 
\begin{eqnarray}
 \frac{2*| t_2 - t_p |}{| t_1 - t_p | + |t_2-t_p|} \mathbb{E}_{p_{data}}[| G(x_{t_1},x_{t_2},t_p) -  x_{t_1} |_1] +&& \\ 
 \frac{2*| t_1 - t_p |}{| t_1 - t_p | + |t_2-t_p|} \mathbb{E}_{p_{data}}[| G(x_{t_1},x_{t_2},t_p) - x_{t_2} |_1],&&\nonumber
\end{eqnarray} 
to better fit generalized synthesis tasks.
We adopt the same settings for the comparison, thus each alternative is viewing the same amount of data at an epoch. 
We note that each epoch using transitive consistency loss, 2) or 3), takes around 2x computational time comparing to those without using transitive properties.
As shown in Figure~\ref{fig:tran_comp}(a), the methods by enforcing transitive properties, both using 2) and 3), converge to better solutions comparing to those without transitive properties.
The network using transitive properties~\eqref{eq:tran_loss2} performs slightly better in convergence, and we use it for the experiments in Section~\ref{sec:experiments}.
Without transitive loss, the images and time ratio of frame synthesis in each batch are randomly selected and could easily bias the training when the batch number is small and constrained by the memory size. 
Thus, the method using 1) oscillates heavily at the beginning and does not converge to a good minimum. 
The network with temporal TV does not perform well as those using transitive properties.
The reason could be that ground-truth frames are not favored by temporal TV in these tasks, especially when large appearance variation happens in videos.
We note that it is possible to apply this loss to other frame interpolation/extrapolation networks to enable them for generic synthesis tasks.


%


\section{Experimental Results}
\label{sec:experiments}
In this section, we compare the proposed MSFSN with
several state-of-the-art methods on benchmark datasets.
We present the quantitative and qualitative comparison in terms of interpolation and extrapolation. 
Finally, we discuss the limitation of the proposed method.
%

\subsection{Interpolation}
We compare our approach against
several methods, including optical flow
techniques: EpicFlow~\cite{revaud-cvpr2015-epicflow} and FlowNet2~\cite{ilg-arxiv-2016-flownet}; and frame synthesis methods: BeyondMSE~\cite{mathieu-arxiv2015-deep}, DVF~\cite{liu-arxiv2017-videoframe} and AdapSC~\cite{niklaus-iccv2017-video}.
We carry out extensive experiments
on public benchmark datasets: UCF-101~\cite{soomro-arxiv2012-ucf101} and THUMOS-15~\cite{idrees-cviu2017-thumos}. UCF-101 and THUMOS-15 contain videos with object motions in relatively low resolution.
%
 %
To synthesize the interpolated
images given the estimated flow fields, we apply the interpolation
algorithm used in the Middlebury interpolation
benchmark~\cite{baker-ijcv2011-database}. 

\begin{figure*}[tb]
\begin{center}
\begin{tabular}{ccc}
\includegraphics[width = 0.21\linewidth]{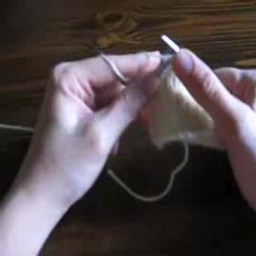}  & \hspace{-4mm}
\includegraphics[width = 0.21\linewidth]{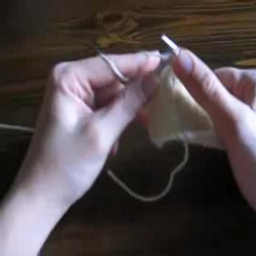}  & \hspace{-4mm}
\includegraphics[width = 0.21\linewidth]{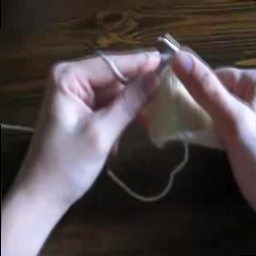}  \\
\includegraphics[width = 0.21\linewidth]{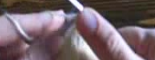}  & \hspace{-4mm}
\includegraphics[width = 0.21\linewidth]{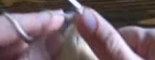}  & \hspace{-4mm}
\includegraphics[width = 0.21\linewidth]{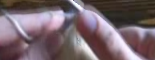}  \\
Ground truth &  EpicFlow &  FlowNet2 \\
\includegraphics[width = 0.21\linewidth]{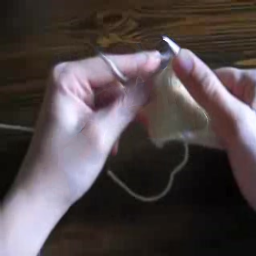}  & \hspace{-4mm}
\includegraphics[width = 0.21\linewidth]{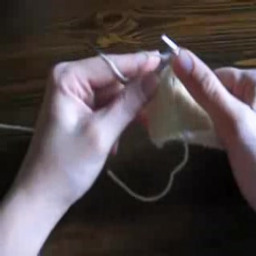}  & \hspace{-4mm}
\includegraphics[width = 0.21\linewidth]{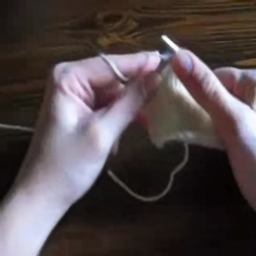}  \\
\includegraphics[width = 0.21\linewidth]{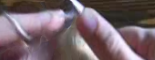}  & \hspace{-4mm}
\includegraphics[width = 0.21\linewidth]{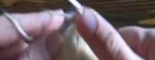}  & \hspace{-4mm}
\includegraphics[width = 0.21\linewidth]{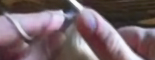}  \\
 DVF &  AdapSC &  Ours \\
\includegraphics[width = 0.21\linewidth]{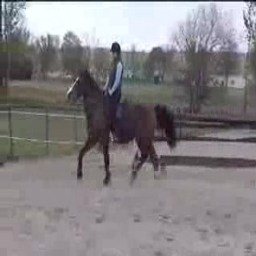}  & \hspace{-4mm}
\includegraphics[width = 0.21\linewidth]{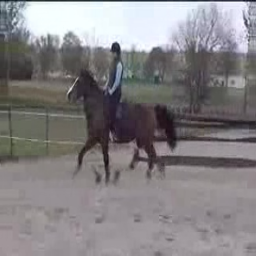}  & \hspace{-4mm}
\includegraphics[width = 0.21\linewidth]{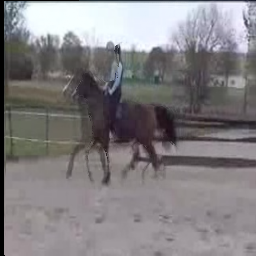}  \\
\includegraphics[width = 0.21\linewidth]{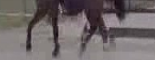}  & \hspace{-4mm}
\includegraphics[width = 0.21\linewidth]{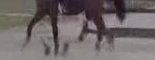}  & \hspace{-4mm}
\includegraphics[width = 0.21\linewidth]{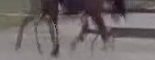}  \\
Ground truth &  EpicFlow &  FlowNet2 \\
\includegraphics[width = 0.21\linewidth]{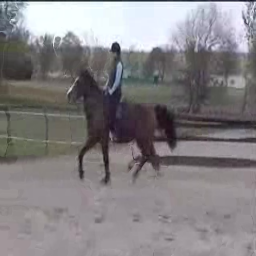}  & \hspace{-4mm}
\includegraphics[width = 0.21\linewidth]{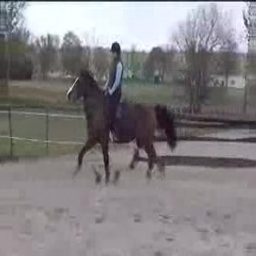}  & \hspace{-4mm}
\includegraphics[width = 0.21\linewidth]{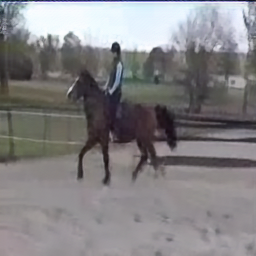}  \\
\includegraphics[width = 0.21\linewidth]{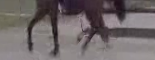}  & \hspace{-4mm}
\includegraphics[width = 0.21\linewidth]{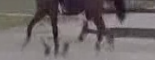}  & \hspace{-4mm}
\includegraphics[width = 0.21\linewidth]{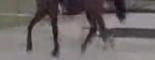}  \\
 DVF &  AdapSC &  Ours \\
\end{tabular}
\caption{Examples of video frame interpolation from UCF-101 dataset~\cite{soomro-arxiv2012-ucf101}. We compare with several state-of-the-art methods, including EpicFlow~\cite{revaud-cvpr2015-epicflow}, FlowNet2~\cite{ilg-arxiv-2016-flownet}, DVF~\cite{liu-arxiv2017-videoframe} and AdapSC~\cite{niklaus-iccv2017-video}.
}
\label{fig:inter_example1}
\end{center}
\end{figure*}

As shown in Figure~\ref{fig:inter_example1}, the correspondence-based methods would generate images with ringing or blurry artifact on the regions where correspondence estimation fails, e.g., front of motor boat, while our network is able to render pleasing results.
We present quantitative comparisons on the benchmark datasets in Table~\ref{tab:perf-interpolation}.
Our approach performs better than flow-based methods EpicFlow~\cite{revaud-cvpr2015-epicflow} and FlowNet2~\cite{ilg-arxiv-2016-flownet}, and comparable to frame synthesis network DVF~\cite{liu-arxiv2017-videoframe}.
The method AdapSC~\cite{niklaus-iccv2017-video} usually generate sharper results than our results, due to their correspondence nature via local filtering.
That is, when the correspondences/spatially-varying kernels are accurately estimated, they can generate sharp results as the input images.

\begin{table}
\centering
\begin{tabular}{l|l|l|ll}
\hline
\multirow{2}{*}{Method} & \multicolumn{2}{c|}{UCF-101}                          & \multicolumn{2}{c}{THUMOS-15}                        \\ \cline{2-5} 
                        & \multicolumn{1}{c|}{PSNR} & \multicolumn{1}{c|}{SSIM} & \multicolumn{1}{c|}{PSNR} & \multicolumn{1}{c}{SSIM} \\ \hline
BeyondMSE               & 32.8                         & 0.93                         & \multicolumn{1}{l|}{32.3}    & 0.91                        \\
EpicFlow                & 34.2                         & 0.95                         & \multicolumn{1}{l|}{33.9}    & 0.94                        \\
FlowNet2                & 34.0                         & 0.94                         & \multicolumn{1}{l|}{33.8}    & 0.94                        \\
DVF                     & 35.8                         & 0.95                         & \multicolumn{1}{l|}{35.4}    & 0.95                        \\
AdapSC                 & 36.2                         & 0.95                         & \multicolumn{1}{l|}{36.4}    & 0.96                        \\
Ours-Gen               & 36.0                         & 0.95                         & \multicolumn{1}{l|}{35.5}    & 0.95                       \\
Ours                    & 35.8                         & 0.95                         & \multicolumn{1}{l|}{35.2}    & 0.95                       \\
\end{tabular}
\caption{Performance (PSNR and SSIM) of video frame interpolation on UCF-101 and THUMOS-15 datasets. }
\label{tab:perf-interpolation}
\end{table}

\begin{table}
\centering
\begin{tabular}{l|l|l|ll}
\hline
\multirow{2}{*}{Method} & \multicolumn{2}{c|}{UCF-101}                          & \multicolumn{2}{c}{THUMOS-15}                        \\ \cline{2-5} 
                        & \multicolumn{1}{c|}{PSNR} & \multicolumn{1}{c|}{SSIM} & \multicolumn{1}{c|}{PSNR} & \multicolumn{1}{c}{SSIM} \\ \hline
BeyondMSE           & 30.6                         & 0.90                         & \multicolumn{1}{l|}{30.2}    & 0.89                        \\
EpicFlow                & 31.3                         & 0.92                         & \multicolumn{1}{l|}{31.0}    & 0.92                        \\
FlowNet2                & 31.8                         & 0.92                         & \multicolumn{1}{l|}{31.7}    & 0.92                        \\
DVF                        & 32.7                         & 0.93                         & \multicolumn{1}{l|}{32.2}    & 0.92                        \\
Ours-Gen               & 32.8                         & 0.93                  & \multicolumn{1}{l|}{32.2}    & 0.92                       \\
Ours                       & 32.4                         & 0.93                         & \multicolumn{1}{l|}{31.9}    & 0.92                       \\
\end{tabular}
\caption{Performance (PSNR and SSIM) of video frame prediction on UCF-101 and THUMOS-15 datasets. }
\label{tab:perf-prediction}
\end{table}


\textbf{Model size}
We include comparisons with other frame-interpolation networks on the number of parameters in Table~\ref{fig:param}.
Unlike the traditional encoder-decoder structure, the proposed network enables parameter sharing as each sub-network level share the same purpose.
The model compression would help reducing memory for model storage during inference, without losing much accuracies, and it could be useful for smartphone applications and edge-device processing in IoT applications.\\

\begin{figure*}[tb]
\begin{center}
\begin{tabular}{cccc}
\includegraphics[width = 0.24\linewidth]{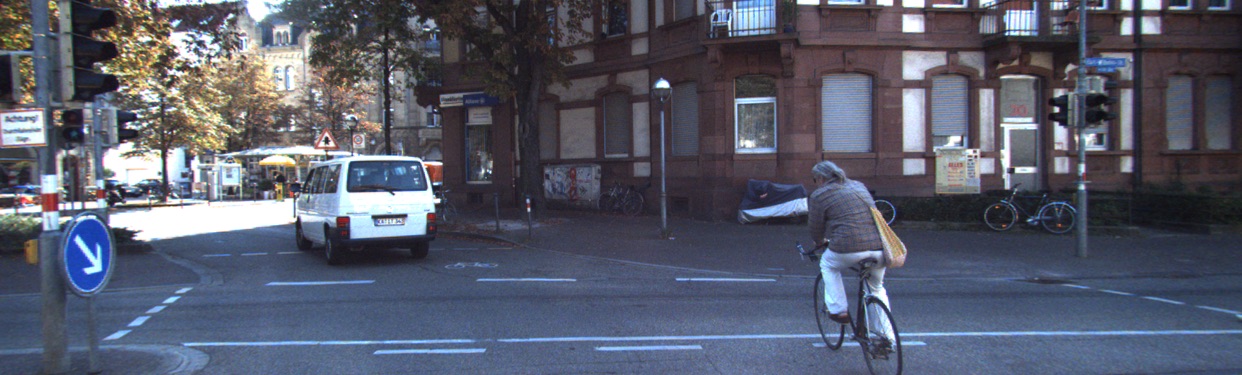}  & \hspace{-3mm}
\includegraphics[width = 0.24\linewidth]{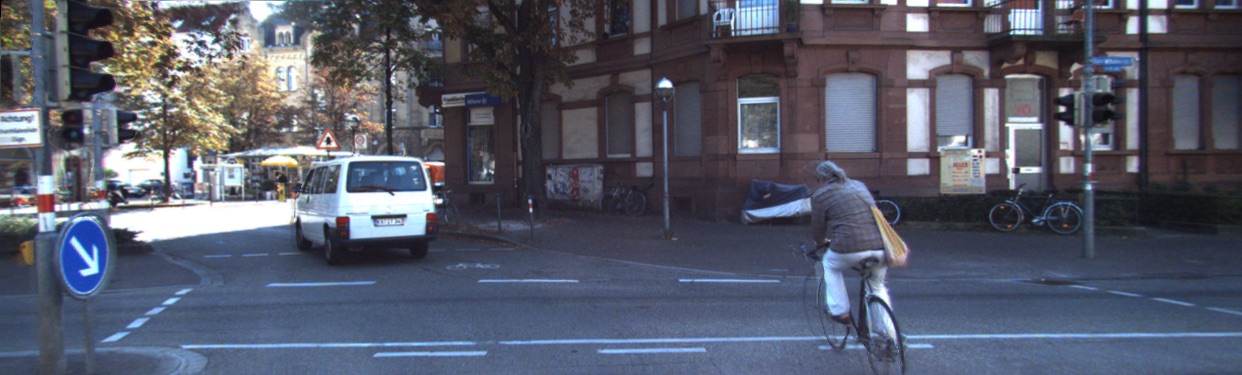}  & \hspace{-3mm}
\includegraphics[width = 0.24\linewidth]{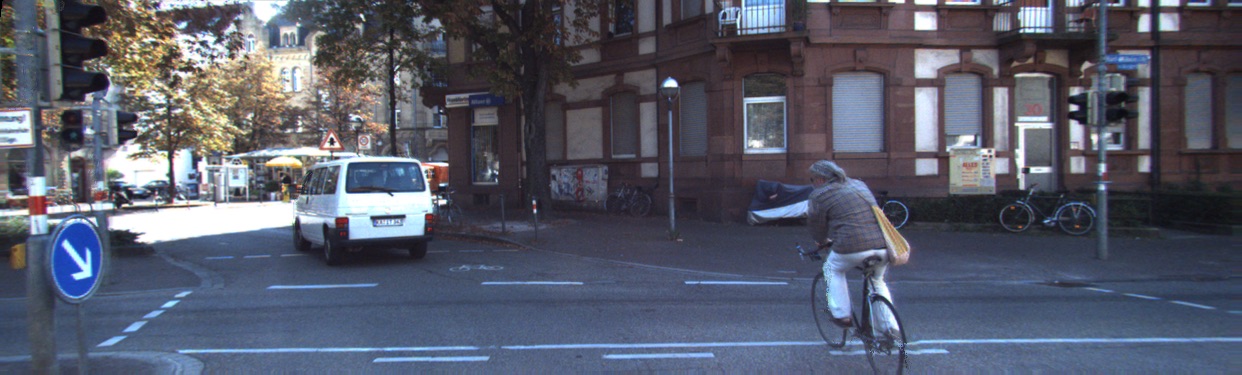}  & \hspace{-3mm}
\includegraphics[width = 0.24\linewidth]{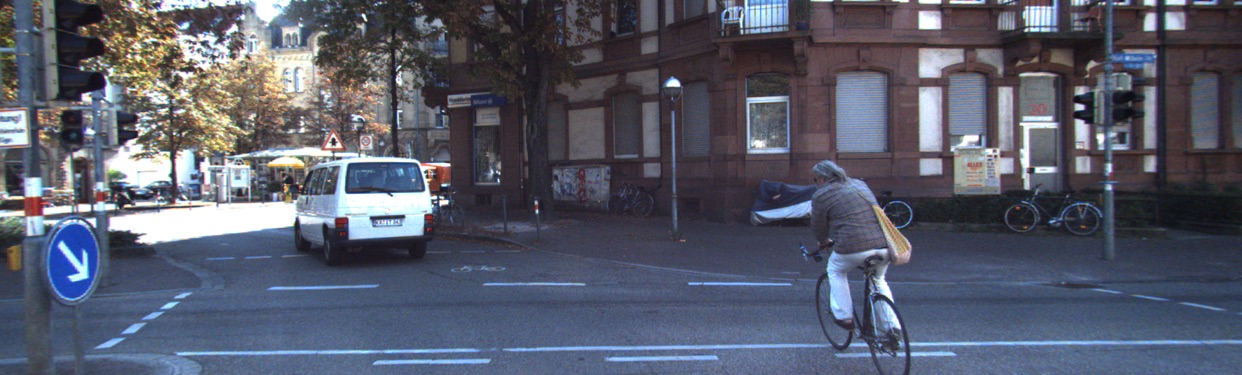}  \\
\includegraphics[width = 0.24\linewidth]{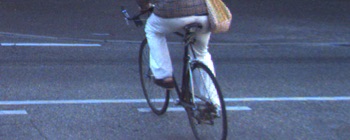}  & \hspace{-3mm}
\includegraphics[width = 0.24\linewidth]{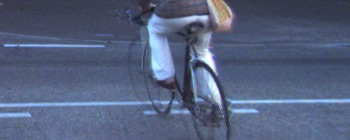}  & \hspace{-3mm}
\includegraphics[width = 0.24\linewidth]{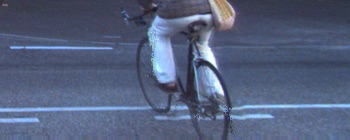}  & \hspace{-3mm}
\includegraphics[width = 0.24\linewidth]{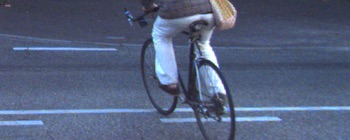}  \\

\includegraphics[width = 0.24\linewidth]{./figs/multi/2/frame_00.jpeg}  & \hspace{-3mm}
\includegraphics[width = 0.24\linewidth]{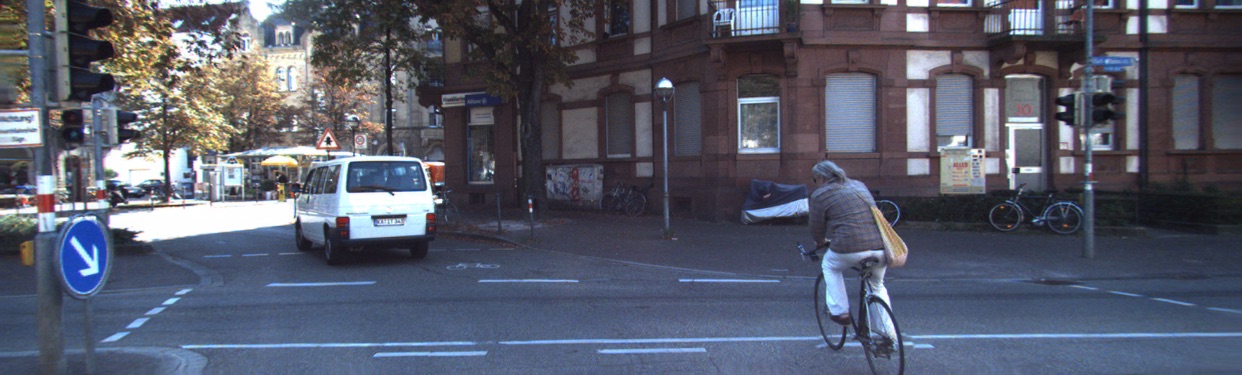}  & \hspace{-3mm}
\includegraphics[width = 0.24\linewidth]{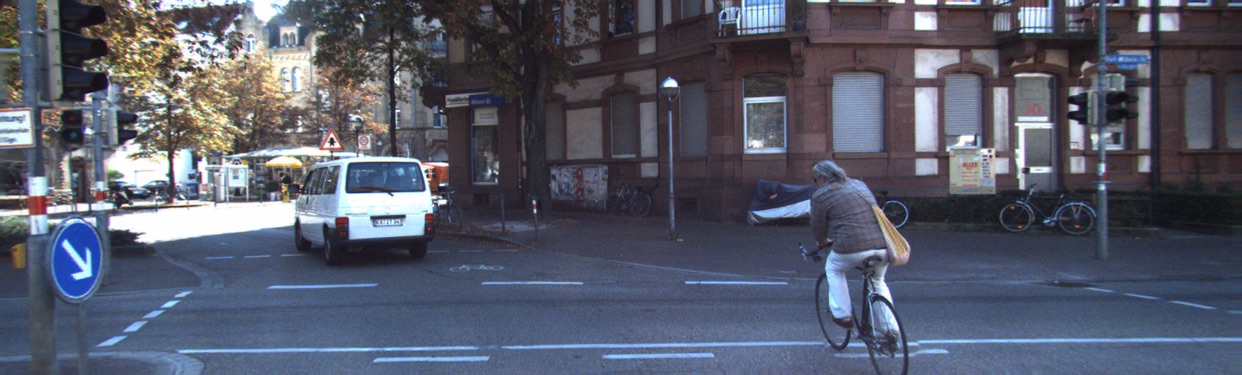}  & \hspace{-3mm}
\includegraphics[width = 0.24\linewidth]{./figs/multi/2/frame_01.jpeg}  \\
\includegraphics[width = 0.24\linewidth]{./figs/multi/2/frame_00_patch.jpeg}  & \hspace{-3mm}
\includegraphics[width = 0.24\linewidth]{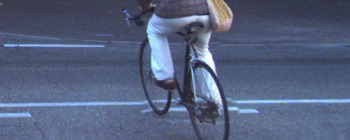}  & \hspace{-3mm}
\includegraphics[width = 0.24\linewidth]{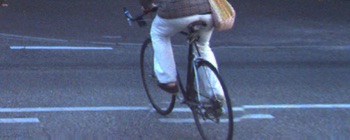}  & \hspace{-3mm}
\includegraphics[width = 0.24\linewidth]{./figs/multi/2/frame_01_patch.jpeg}  \\

\includegraphics[width = 0.24\linewidth]{./figs/multi/2/frame_00.jpeg}  & \hspace{-3mm}
\includegraphics[width = 0.24\linewidth]{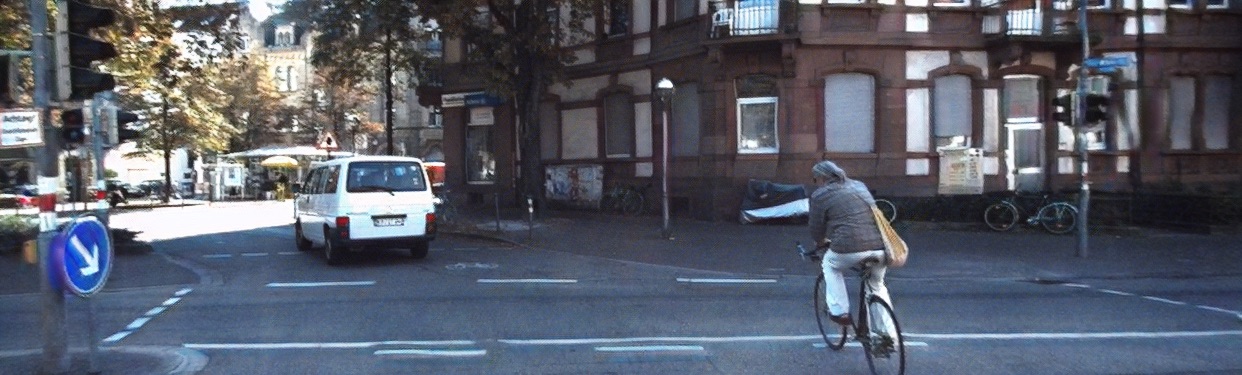}  & \hspace{-3mm}
\includegraphics[width = 0.24\linewidth]{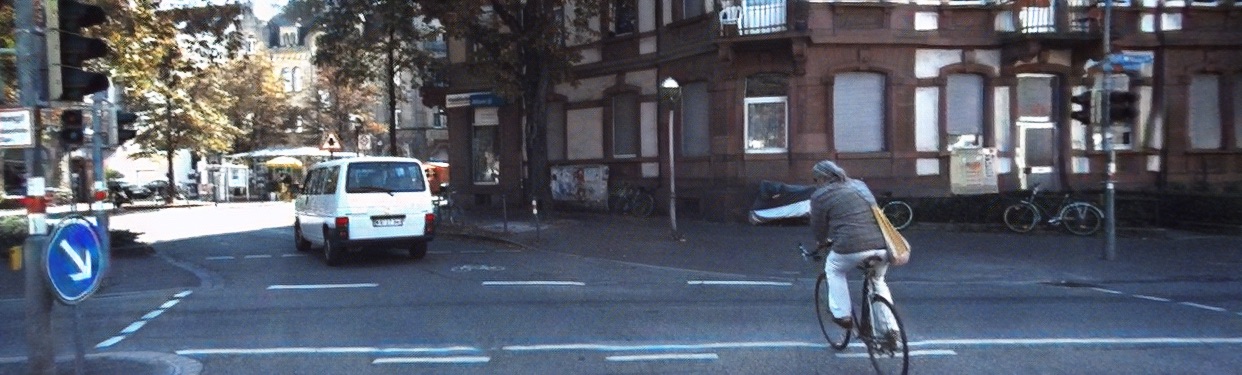}  & \hspace{-3mm}
\includegraphics[width = 0.24\linewidth]{./figs/multi/2/frame_01.jpeg}  \\
\includegraphics[width = 0.24\linewidth]{./figs/multi/2/frame_00_patch.jpeg}  & \hspace{-3mm}
\includegraphics[width = 0.24\linewidth]{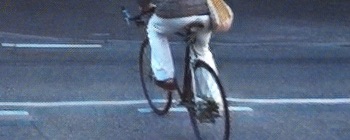}  & \hspace{-3mm}
\includegraphics[width = 0.24\linewidth]{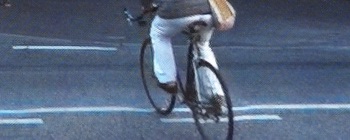}  & \hspace{-3mm}
\includegraphics[width = 0.24\linewidth]{./figs/multi/2/frame_01_patch.jpeg}  \\

(a) First input frame & (b) Interpolated frame 1 & (c) Interpolated frame 3 & (d) Second input frame \\
\end{tabular}
\caption{Example of multiple frame interpolation from KITTI dataset~\cite{Menze2015CVPR}. For this example, we interpolate three equally spaced frames in-between and show comparison with FlowNet2~\cite{ilg-arxiv-2016-flownet} and AdapSC~\cite{niklaus-iccv2017-video}.
The first to third rows are results from FlowNet2, AdapSC and ours respectively. Here we show first and third interpolated frames for comparison.
}
\label{fig:multi_example1}
\end{center}
\end{figure*}

\textbf{Interpolating multiple frames}
Unlike those networks trained on specific interpolation settings, e.g., interpolating middle in-between frame, our method is capable of synthesizing frames at any in-between position, without computing correspondences.
We show comparison with flow-based method~\cite{ilg-arxiv-2016-flownet} and frame interpolation network~\cite{niklaus-iccv2017-video} on interpolating multiple in-between frames.
To compare with the network~\cite{niklaus-iccv2017-video} which can only interpolate middle in-between frame, we evaluate on the scenario of interpolating three equally spaced frames. 
In this case, we obtain their results by two-stage interpolation, which is first interpolating middle frame and then interpolating the other two using the middle frame as an input.
As shown in Figure~\ref{fig:multi_example1}, our method renders sharp results while maintains the straight road lines while others cannot.

\textbf{User study}
To better understand the visual quality of different methods, we conducted a user study on the interpolated frames, comparing with state-of-the-art methods, FlowNet2~\cite{ilg-arxiv-2016-flownet}, EpicFlow~\cite{revaud-cvpr2015-epicflow}, DVF~\cite{liu-arxiv2017-videoframe}, AdapSC~\cite{niklaus-iccv2017-video} and the ground truth (GT). 
We develop a web-based system to display and collect study results. 
The system provides two side-by-side results at a time, one from our method and the other from a randomly selected method.
Each pair of results is randomly selected and placed. 
There are 42 subjects participated in the user study and each of them was asked to select on 20 pairs of comparison.
As shown in Figure~\ref{fig:tran_comp}(b), our method is preferred over FlowNet2, EpicFlow and DVF, and obtain comparable results comparing with AdapSC. 

\subsection{Extrapolation}
\begin{figure*}[t]
\begin{center}
\begin{tabular}{ccc}
\includegraphics[width = 0.21\linewidth]{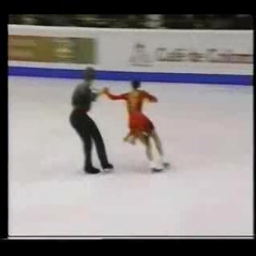}  & \hspace{-4mm}
\includegraphics[width = 0.21\linewidth]{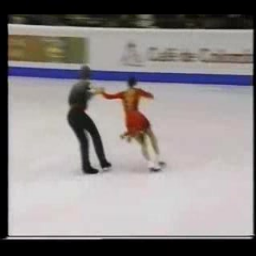}  & \hspace{-4mm}
\includegraphics[width = 0.21\linewidth]{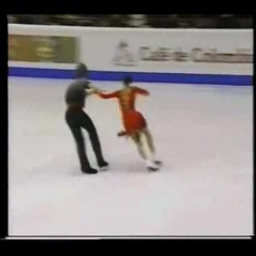}  \\
\includegraphics[width = 0.21\linewidth]{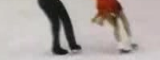}  & \hspace{-4mm}
\includegraphics[width = 0.21\linewidth]{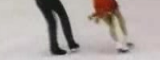}  & \hspace{-4mm}
\includegraphics[width = 0.21\linewidth]{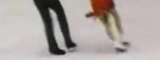}  \\
Input frame 1 &  Input frame 2 &  Ground truth\\
&\hspace{-4mm}
\includegraphics[width = 0.21\linewidth]{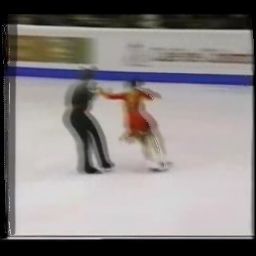}  & \hspace{-4mm}
\includegraphics[width = 0.21\linewidth]{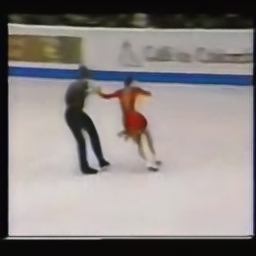}  \\
&\hspace{-4mm}
\includegraphics[width = 0.21\linewidth]{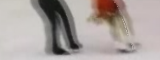}  & \hspace{-4mm}
\includegraphics[width = 0.21\linewidth]{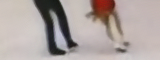}  \\
& FlowNet2 &  Ours\\

\includegraphics[width = 0.21\linewidth]{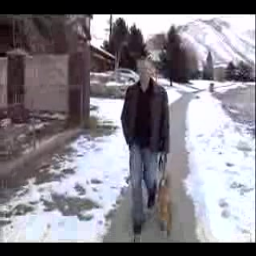}  & \hspace{-4mm}
\includegraphics[width = 0.21\linewidth]{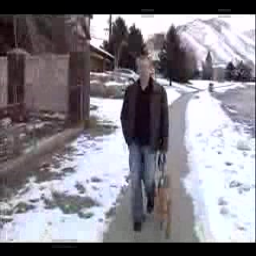}  & \hspace{-4mm}
\includegraphics[width = 0.21\linewidth]{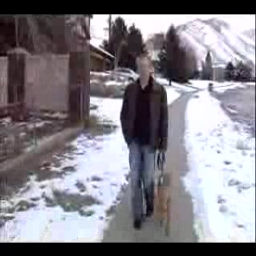}  \\
\includegraphics[width = 0.21\linewidth]{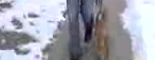}  & \hspace{-4mm}
\includegraphics[width = 0.21\linewidth]{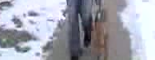}  & \hspace{-4mm}
\includegraphics[width = 0.21\linewidth]{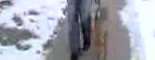}  \\
Input frame 1 &  Input frame 2 &  Ground truth\\
&\hspace{-4mm}
\includegraphics[width = 0.21\linewidth]{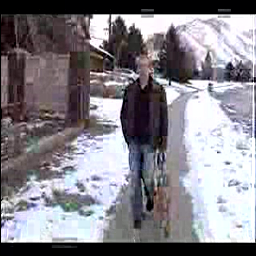}  & \hspace{-4mm}
\includegraphics[width = 0.21\linewidth]{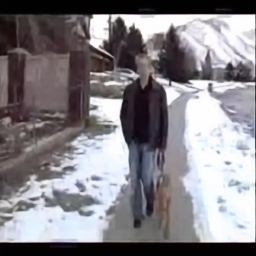}  \\
&\hspace{-4mm}
\includegraphics[width = 0.21\linewidth]{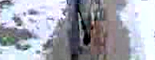}  & \hspace{-4mm}
\includegraphics[width = 0.21\linewidth]{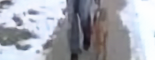}  \\
& FlowNet2 &  Ours\\
\end{tabular}
\caption{Examples of video frame extrapolation from UCF-101 dataset~\cite{soomro-arxiv2012-ucf101}. We compare with FlowNet2~\cite{ilg-arxiv-2016-flownet}.}
\label{fig:extra_example1}
\end{center}
\end{figure*}

The trained model can be directly applied to video extrapolation tasks without fine tuning.
Most of the interpolation networks mentioned are trained for a specific setting, and therefore cannot apply directly to the extrapolation cases. 
We compare with state-of-the-art methods on UC-101 and THUMOS-15 datasets and present quantitative results in Table~\ref{tab:perf-prediction}.
Qualitative comparisons are given in Figure~\ref{fig:extra_example1}. 

\subsection{Limitation and Discussion}
While our network is capable to handle large motion, the results from our method are usually more blurry than than corresponding-based method~\cite{niklaus-iccv2017-video}, as they fuse correponding pixels through a local window filtering. 
We will explore the framework to combine the benefit of corresponding-based method into the unified framework in our future work.
Another limitation of the proposed network is that the synthesized frame is assumed to follow the motion momentum, which is constrained and parameterized using one temporal variable. 
How to generalize the motion patterns in the network will also be an interesting research direction.

\section{Conclusion}
In this paper, we propose a unified deep neural network for video frame synthesis. 
The proposed model progressively predicts interpolated/extrapolated frames in a coarse-to-fine manner. 
We introduce a transitive consistency loss to facilitate the network training and enable the network for both interpolation and extrapolation capabilities.
By sharing parameters across pyramid levels, the network is compact and is practical to use in devices where memory and computation power are limited.
The proposed model can be easily extend to scenarios, e.g., long-range motion, where deeper pyramid is needed.
Extensive evaluations on benchmark datasets demonstrate that the proposed model performs favorably against state-of-the-art algorithms.

{\small
\bibliographystyle{ieee}
\bibliography{frameinterpolation}
}

\end{document}